%% file: main.tex
\documentclass[journal]{IEEEtran}

\input{math.tex}

\usepackage{kotex}
\usepackage{cite}
\usepackage[square, sort,comma,numbers]{natbib}
\usepackage[pdftex]{graphicx}
\usepackage{xcolor}
\usepackage{tikz}
\usepackage{mathtools}
\usepackage{amsmath}
\usepackage{amsfonts}
\usepackage{bm}

\usepackage{array}
\usepackage{multirow}
\usepackage{booktabs}
\usepackage{url}
\usepackage{hyperref}
\usepackage[ruled,linesnumbered,lined,commentsnumbered]{algorithm2e}
\usepackage{algpseudocode}
\usepackage{stfloats}

\newcommand{\eg}{{\em e.g.}}           
\newcommand{\ie}{{\em i.e.}}           
\newcommand{\etc}{{\em etc.}}          

\newcommand{\revise} {\color{black}}

\begin{document}
\title{Uncertainty-Aware Variational-Recurrent Imputation Network for Clinical Time Series}

\author{Ahmad Wisnu Mulyadi, Eunji Jun, and~Heung-Il~Suk,~\IEEEmembership{Member,~IEEE}
\thanks{A. W. Mulyadi and E. Jun are with the Department of Brain and Cognitive Engineering, Korea University, Seoul 02841, Korea (e-mail: wisnumulyadi@korea.ac.kr, ejjun92@korea.ac.kr). H.-I. Suk is with the Department of Artificial Intelligence and the Department of Brain and Cognitive Engineering, Korea University, Seoul 02841 Korea (e-mail: hisuk@korea.ac.kr). (\textit{Corresponding author: Heung-Il Suk})}}

\markboth{UNDER REVIEW}%
{Mulyadi \MakeLowercase{\textit{et al.}}: Uncertainty-Aware Variational-Recurrent Imputation Network for Clinical Time Series}

\maketitle

\begin{abstract}
Electronic health records (EHR) consist of longitudinal clinical observations portrayed with sparsity, irregularity, and high-dimensionality, which become major obstacles in drawing reliable downstream clinical outcomes. Although there exist great numbers of imputation methods to tackle these issues, most of them ignore correlated features, temporal dynamics and entirely set aside the uncertainty. Since the missing value estimates involve the risk of being inaccurate, it is appropriate for the method to handle the less certain information differently than the reliable data. In that regard, we can use the uncertainties in estimating the missing values as the fidelity score to be further utilized to alleviate the risk of biased missing value estimates. In this work, we propose a novel variational-recurrent imputation network, which unifies an imputation and a prediction network by taking into account the correlated features, temporal dynamics, as well as the uncertainty. Specifically, we leverage the deep generative model in the imputation, which is based on the distribution among variables, and a recurrent imputation network to exploit the temporal relations, in conjunction with utilization of the uncertainty. We validated the effectiveness of our proposed model on two publicly available real-world EHR datasets: PhysioNet Challenge 2012 and MIMIC-III, and compared the results with other competing state-of-the-art methods in the literature.
\end{abstract}


\IEEEpeerreviewmaketitle
\section{Introduction}
\IEEEPARstart{E}{lectronic} health records (EHRs) store longitudinal data consisting of patients' clinical observations in the intensive care unit (ICU). Despite the surge of interest in clinical research on EHR, it still holds diverse challenging issues to be tackled, these include high-dimensionality, temporality, sparsity, irregularity, and bias \cite{cheng2016,yadav2018}. Specifically, sequences of multidimensional medical measurements are recorded irregularly in terms of its variables and times. The reasons behind these typical measurements are diverse, such as lack of collection, documentation, or even recording faults \cite{wells2013,cheng2016}. Since it carries essential information regarding a patient's health status, improper handling of missing values might cause an unintentional bias \cite{wells2013,jones2017}, yielding an unreliable downstream analysis and verdict.

Complete-case analysis is an approach that draws clinical outcomes by disregarding the missing values and relying only on the observed values. However, excluding the missing data yields poor performance at high missing rates and also requires modeling separately for the distinct dataset. In fact, those missing values reflect the decisions made by health-care providers \cite{lipton2016}. Therefore, the missing values and their patterns contain information regarding a patient's health status \cite{lipton2016} and correlate with the outcomes or target labels \cite{che2018}. Thus, we resort to the imputation approach by exploiting those missing patterns to improve the prediction of the clinical outcomes as the downstream task.

There exist numerous strategies for imputing missing values in the literature. Generally, imputation methods can be classified into a deterministic or stochastic approach, depending on the use of randomness \cite{brick1996}. Deterministic methods, such as mean \cite{little1987} and median filling \cite{acuna2004}, produce only one possible value when estimating the missing values. However, it is desirable for the imputation model to generate values or samples by considering the distribution of the available observed data. Thus, it leads us to the employment of  stochastic-based imputation methods. 

The recent rise of deep learning models has offered potential solutions in accommodating such circumstances. Variational autoencoders (VAEs) \cite{kingma2014} and generative adversarial networks (GANs) \cite{goodfellow2014} exploit the latent distribution of high-dimensional incomplete data and generate comparable data points as the approximation estimates for the missing or corrupted values \cite{nazabal2018,luo2018,jun2019}. However, such deep generative models are insufficient for estimating the missing values of multivariate time series owing to their nature of ignoring temporal relations between a span of time points. On the other hand, by virtue of recurrent neural networks (RNNs), which have proved to perform remarkably well in modeling sequential data, we can estimate the complete data by taking into account the temporal characteristics. In this approach, \mbox{GRU-D} \cite{che2018} introduced a modified gated-recurrent unit (GRU) cell to model missing patterns in the form of masking vectors and temporal delays. Likewise, BRITS \cite{cao2018} exploited the temporal relations of bidirectional dynamics by considering feature correlations in estimating the missing values. Moreover, inspired by the residual networks and graph-based methods, \cite{ma2019} proposed the LIME-RNN to tackle the imputation task by integrating the previous hidden states of RNN in the forms of linear memory vector.

Even though such models employed the stochastic approach for inferring and generating samples by utilizing both features and temporal relations, they scarcely exploited the uncertainty in estimating the missing values in multivariate time series data (\ie, since the imputation estimates are not thoroughly accurate, we may introduce a fidelity score denoted by the uncertainty, which enhances the downstream task performance by emphasizing the reliable information more than the less certain information) \cite{he2010,gemmeke2010,jun2019}. We can use the imputation model, which captures the aleatoric uncertainty in estimating the missing values by placing a distribution over the output of the model \cite{kendall2017}. In particular, we would like to estimate the heteroscedastic aleatoric uncertainties, which are useful in cases where observation noises vary with the input \cite{kendall2017}.

In this work, we define our primary task as the prediction of in-hospital mortality on clinical time series data. However, since such data are portrayed with sparse and irregularly-sampled characteristics, we devise an imputation model as the secondary problem to enhance the clinical outcome predictions. We propose a novel variational-recurrent imputation network (V-RIN) which unifies the imputation and prediction networks for multivariate time series EHR data, governing both correlations among variables and temporal relations. Specifically, given the sparse data, an inference network of VAEs is employed to capture data distribution in the latent space. From this, we employ a generative network to obtain the reconstructed data as the imputation estimates for the missing values and the uncertainty indicating the imputation fidelity score. Then, we integrate the temporal and feature correlations into a combined vector and feed it into a novel \emph{uncertainty-aware GRU} in the recurrent imputation network. Finally, we obtain the mortality prediction as a clinical verdict from the complete imputed data. In general, our main contributions in this study are as follows:

\begin{itemize}
    \item {We estimate the missing values by utilizing a deep generative model combined with a recurrent imputation network to capture both feature correlations and the temporal dynamics jointly, yielding the uncertainty.}
    \item {We effectively incorporate the uncertainty with the imputation estimates in our novel uncertainty-aware GRU cell for better prediction results.}
    \item {We evaluate the effectiveness of the proposed models by training the imputation and prediction networks jointly in an end-to-end manner, achieving superior performance on real-world multivariate time series EHR data compared to other competing state-of-the-art methods.}
\end{itemize}

This study extends the preliminary work published in \cite{jun2019}. Unlike to the preceding study, we have further expanded the proposed model by introducing more complex recurrent imputation networks, which utilize the uncertainties, instead of vanilla RNNs. We also include two additional real-world EHR datasets in our experiments, described in Section (\ref{section:experiments_dataset}), and validate the robustness of our proposed networks by comparing them with existing state-of-the-art models in the literature (Section \ref{section:experiments_compmodels}). Furthermore, we have conducted extensive experiments to discover the impacts of missing value estimation by utilizing the uncertainties when performing the downstream task (Section \ref{section:experiments_ablation} - \ref{section:experiments_imputation}).

The rest of the paper is organized as follows. In Section \ref{section:related_work}, we discuss the works closely related to our proposed model in imputing the missing values. In Section \ref{section:methods}, we detail our proposed model. In Section \ref{section:experiments}, we report on the experimental results and analysis by comparing them with state-of-the-art methods. Finally, we conclude the work in Section \ref{section:conclusion}.

\begin{figure*}[t]
\includegraphics[width=1.0\textwidth]{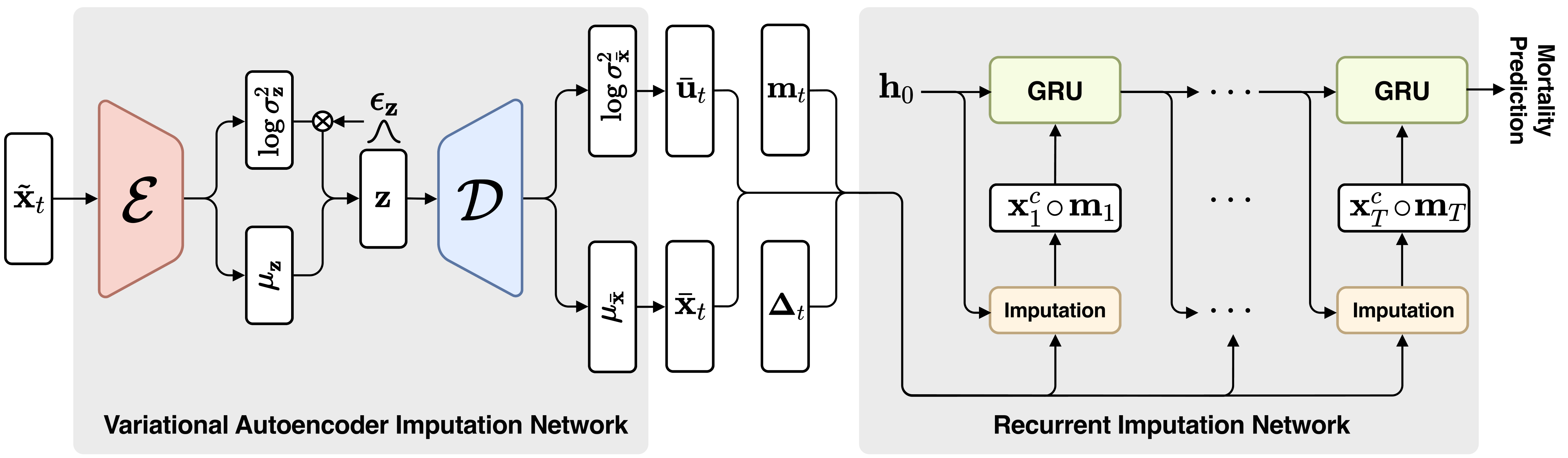}
\caption{\revise Architecture of the proposed model, which unifies two key networks, namely variational autoencoder and recurrent imputation network. The model considers the feature correlations, temporal relations, and the uncertainties in estimating the missing values to get a better prediction of clinical outcome. Refer to Section \ref{section:methods} for more details of the notation.}
\label{fig:architecture}
\end{figure*}

\section{Related Work}
\label{section:related_work}
Imputation strategies have been extensively devised to resolve the issue of sparse high-dimensional time series data by means of statistics, machine learning, or deep learning methods. For instance, previous works exploited statistical attributes of observed data, such as mean \cite{little1987} and median filling \cite{acuna2004}, which clearly ignored the temporal relations and the correlations among variables.  From the machine learning approaches, $k$-nearest neighbors (KNN) \cite{troyanskaya2001}, and principal component analysis (PCA) \cite{mohamed2009} were proposed by considering the relationships of the features either in the original or in the latent space. In addition, imputation methods based on the expectation-maximization (EM) algorithm \cite{dempster1977, chen2019} were also proposed in tackling the missing values estimations. Furthermore, multiple imputation by chained equations (MICE) \cite{white2011,azur2011}  introduced variability by means of repeating the imputation process multiple times. However, these methods ignore the temporal relations as crucial attributes in time series modeling and disregard the exploitation of the uncertainties in estimating the missing values.

The deep learning-based imputation models, are closely related to our proposed models. A previous study \cite{nazabal2018} leveraged VAEs to generate stochastic imputation estimates by exploiting the distribution and correlations of features in the latent space. However, it ignored the temporal relations as well as the uncertainties. Meanwhile, GP-VAE \cite{fortuin2019} was proposed to obtain the latent representation by means of VAEs and model temporal dynamics in the latent space using the Gaussian process. However, since the model is merely focused on the imputation task, they required a separate model for the  further downstream outcome. Recently, the sparse autoencoder (SAE) was exploited to extract the features in combination with a complex recurrent broad learning system (RBLS) \cite{xu2020} to capture the temporal dynamic characteristics in the time series. Nevertheless, uncertainties in imputing the missing values were ignored.

To deal with time series data, a series of RNN-based imputation models were proposed. GRU-D \cite{che2018} considered the temporal dynamics by incorporating the missing patterns, together with the mean imputation, and forward filling with past values using temporal decay factor. Similarly, \mbox{GRU-I} \cite{luo2018} trained the RNNs using such temporal decay factor and further incorporated an adversarial scheme of GANs as the stochastic approach. In the meantime, BRITS \cite{cao2018} were proposed to combine the feature correlations and temporal dynamic networks using bidirectional dynamics, which enhanced the accuracy by estimating missing values in both forward and backward directions. By considering the delayed gradients of the missingness in the forward and backward directions, their models were able to achieve more accurate missing values imputations. Likewise, M-RNN \cite{yoon2017} utilized bidirectional recurrent dynamics by operating interpolation (intra-stream) and imputation (inter-stream). Despite temporal dynamics and stochastic methods being considered in the models, the uncertainties for imputation purposes were scarcely incorporated. 

As we are unsure of the actual values, we argue that these uncertainties are beneficial and can be utilized to estimate the missing values. Such uncertainty can be captured by accommodating the distribution over the model output \cite{kendall2017}. For this purpose, we exploited VAEs \cite{kingma2014} as the Bayesian networks, which are able to model the data distribution. In this work, we introduce the uncertainty as the imputation fidelity of estimates, which compensates for the potential impairment of imputation estimates. Therefore, we assumed that our model could provide reliable estimates while giving less attention to the unreliable ones determined by its uncertainties. We expect to obtain better estimates of the missing values leading to a better prediction performance as a downstream task. However, since VAEs alone are not designed to model the temporal dynamics, we combined the model with the recurrent imputation network by further utilizing those uncertainties. Thus, our proposed model differs from the aforementioned models in which that it integrates the imputation and prediction networks jointly, and the utilization of the uncertainties serve as the major motivation in our works.


\section{Proposed Methods}
\label{section:methods}
Our proposed model architecture consists of two key networks -- an imputation and a prediction network -- as depicted in Fig. \ref{fig:architecture}. The imputation network is devised based on VAEs to capture the latent distribution of the sparse data by means of its inference network (\ie, encoder $\mathcal{E}$). Then, the subsequent generative network of VAEs (\ie, decoder $\mathcal{D}$) estimates the reconstructed data distribution. We regard its reconstructed values as the imputation estimates while exploiting its variances as the uncertainties to be further utilized in the recurrent imputation network sequentially.

The succeeding recurrent imputation networks are built upon RNNs to model the temporal dynamics. For each time step, we employ the regression layer to model the temporal relations incorporated within the hidden states of RNNs cell in imputing the missing values. However, as we also consider the time gap between observed values, we incorporate the time decay factor, which is then exploited in such hidden states, leading to \textit{decayed hidden states}. Eventually, by systematically unifying the imputation estimates obtained from VAEs and RNNs, we expect to acquire more likely estimates by considering feature correlations and temporal relations over time, including the utilization of the uncertainty. {\revise By doing so, we expect a more reliable prediction outcomes (\ie \ in-hospital mortality) could be obtained.} We describe each of the networks more specifically in the following sections after introducing the data representation.

\subsection{Data Representation}

Given the multivariate time series EHR data of $N$ number of patients, a set of clinical observations and their corresponding label is denoted as $\{\rmX^{(n)}, y^{(n)}\}^{N}_{n=1}$. For each patient, we have $\rmX^{(n)} = [\rvx^{(n)}_1, \ldots, \rvx^{(n)}_t, \ldots, \rvx^{(n)}_T]^\top \withdim{T \times D}$, where $T$ and $D$ represent the time points and variables, respectively; $\rvx^{(n)}_t \withdim{D}$ denotes all observable variables at time point $t$, and $x^{d,(n)}_t$ is the $d$-th element of variables at time point $t$. In addition, it has the corresponding clinical label $y^{(n)} \in \{0,1\}$, representing the clinical outcome. In our case, it denotes the in-hospital mortality of a patient, which falls into a binary classification problem. For the sake of clarity, we omit the superscript $(n)$ hereafter. 

As $\rmX$ is characterized with sparsity properties, we address the missing values by introducing masking matrix $\rmM \in \{0,1\}^{T\times D}$, indicating whether values are observed or missing. In addition, we define a new data representation $\tilde{\rmX} = [\tilde{\rvx}_1, \ldots, \tilde{\rvx}_t, \ldots, \tilde{\rvx}_T]^\top \withdim{T \times D}$ to be fed into the model, where we initialize the missing value with zero \cite{lipton2016, nazabal2018, jun2019} as follows:
\begin{align*}
    m^d_t &= 
    \begin{cases}
        1 &\quad \textrm{if} \ x^d_t \ \textrm{is observed},\\
        0 &\quad \textrm{otherwise}
    \end{cases},
    \tilde{x}^d_t &= 
    \begin{cases}
        x^d_t &\quad \textrm{if} \ m^d_t = 1,\\
        0 &\quad \textrm{otherwise.}
    \end{cases}
\end{align*}
Another consideration in dealing with the sparse data is that there exists a time gap between two observed values. Such information in fact carries a piece of essential information in estimating the missing values over the times. Thus, we accommodate this information by further devising the time delay matrix ${\mDelta} \withdim{T\times D}$, which is derived from  $\rvs \withdim{T}$, denoting the timestamp of the measurement. As initialization, we fix ${\vdelta}_t = \mOne$ for the $t=1$, while setting the time delay  for the remaining times $(t > 1)$ by referring to a masking matrix ${\revise \rmM}$ and a timestamp vector ${\revise \rvs}$ as follows:
\begin{align*}
    \Delta^d_t = 
    \begin{cases}
        s_t - s_{t-1} &\quad \textrm{if} \ m^d_{t-1}=1, \\
        s_t - s_{t-1} + \Delta^d_{t-1} &\quad \textrm{otherwise.} \\
    \end{cases}
\end{align*}


\begin{figure*}[t]
\centering
\includegraphics[width=0.8\linewidth]{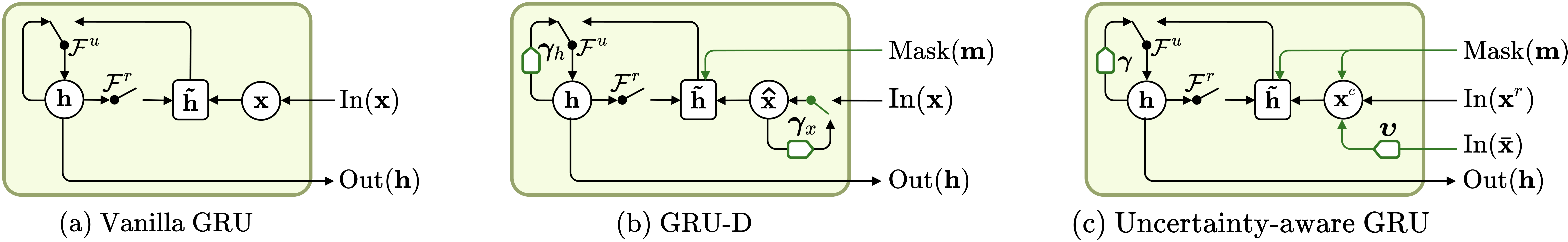}
\caption{Graphical illustrations of comparing three different forms of GRU cells ($\gF^u$: update gate, $\gF^r$: reset gate).}
\label{fig:grus_cells}
\end{figure*}


\subsection{VAE-based Imputation Network}
Given the observations at each time point $\tilde{\rvx}_t$, we infer  $\rvz\withdim{k}$ as the latent representation with $k$ as its corresponding dimension by making use of the inference network, using the true posterior distribution $p_\phi(\rvz|\tilde{\rvx}_t)$. Intuitively, we assume that $\tilde{\rvx}_t$ is generated from some unobserved random variable $\rvz$ by some conditional distribution $p_\theta(\tilde{\rvx}_t|\rvz)$, while $\rvz$ is generated from a prior distribution $p_\theta(\rvz)$, which can be interpreted as the hidden health status of the patient. In addition, we define the marginal likelihood as $p_\theta(\tilde{\rvx}_t)$ by integrating out the joint distribution $p_\theta(\tilde{\rvx}_t,\rvz)$ for $\rvz$ defined as

\begin{equation*}
     p_\theta(\tilde{\rvx}_t) = \int p_\theta(\tilde{\rvx}_t,\rvz)d\rvz = \int p_\theta(\tilde{\rvx}_t|\rvz) p_\theta(\rvz) d\rvz.
\end{equation*}

\noindent However, in practice, this is analytically intractable. Since, $p_\theta(\rvz|\tilde{\rvx}_t) = p_\theta(\tilde{\rvx}_t|\rvz)p_\theta(\rvz)/p_\theta(\tilde{\rvx}_t)$, the true posterior becomes intractable as well. Therefore, {\revise we approximate it with $q_\phi(\rvz|\tilde{\rvx}_t)$ using a Gaussian distribution $\mathcal{N}(\vmu_\rvz, \textrm{diag}(\vsigma^2_\rvz))$  \cite{jun2019, reyes2018}}, where the mean and log-variance are obtained such that
\begin{align*}
    \vmu_{\revise \rvz} &= \gE_{\revise \vmu}(\tilde{\rvx}_t;\phi), \quad \log \vsigma^2_{\revise \rvz} = \gE_\vsigma(\tilde{\rvx}_t;\phi),
\end{align*}
\noindent where $\gE_{\{{\revise \vmu, \vsigma} \}}$ denotes the inference network with parameter $\phi$. Furthermore, we apply the reparameterization trick  \cite{kingma2014} as $\rvz = \vmu_{\revise \rvz} + \vsigma_{\revise \rvz} \odot \ \vepsilon_{\revise \rvz}$, where $\vepsilon_{\revise \rvz} \sim \mathcal{N}(\mZero,\mIdentity)$, with $\odot$ denoting element-wise multiplication, thus, making it possible to be differentiated and trained using standard gradient methods. 

Furthermore, given this latent vector $\rvz$, we estimate $p_\theta(\tilde{\rvx}_t|\rvz)$ by means of the generative network $\gD$ with parameters $\theta$ as 
\begin{align*}
    \vmu_{{\revise \bar{\rvx}},t} &= \gD_{\revise \vmu}(\rvz;\theta), \quad \log \vsigma^2_{{\revise \bar{\rvx}},t} = \gD_{\revise \vsigma}(\rvz;\theta),
\end{align*}
\noindent where $\vmu_{{\revise \bar{\rvx}},t}$ and $\vsigma^2_{{\revise \bar{\rvx}},t}$ denote the mean and variance of reconstructed data distribution, respectively. We regard the mean as the estimate to the missing values and maintain the observed values in $\bar{\rvx}_t$ by making use of corresponding masks vector as
\begin{align*}
    \bar{\rvx}_t = \rvm_t \odot \tilde{\rvx}_t + (\mOne - \rvm_t) \odot \vmu_{{\revise \bar{\rvx}},t}.
\end{align*}
\noindent In the meantime, we regard the variance of reconstructed data as the uncertainty to be further utilized in the recurrent imputation process. For this purpose, we introduce an uncertainty matrix $\bar{\rmU} \withdim{T\times D}$ with $\hat{\mSigma}_{\revise \bar{\rvx}}= [\textrm{diag}(\vsigma_{{\revise \bar{\rvx}},1}), \ldots, \textrm{diag}(\vsigma_{{\revise \bar{\rvx}},t}), \ldots, \textrm{diag}(\vsigma_{{\revise \bar{\rvx}},T})]^\top \withdim{T \times D}$. We quantify this uncertainty as the fidelity score of the missing value estimates. In particular, we set the corresponding uncertainty to zero if the data is observed, indicating that we are confident with full trust in the observation, and set it as a value $\sigma^d_{{\revise \bar{\rvx}},t}$ if the corresponding value is missing as
\begin{align*}
    \bar{\rvu}_t = (\mOne - \rvm_t) \odot \hat{\vsigma}_{{\revise \bar{\rvx}},t}
\end{align*}

Finally, as an output of this VAE-based imputation network, we acquire the set $\{\bar{\rmX}, \bar{\rmU}\}$ denoting the imputed values and their corresponding uncertainty, respectively. Furthermore, to alleviate the bias of missing value estimations, we utilize this uncertainty in the following recurrent imputation network. 


\subsection{Recurrent Imputation Network}
The recurrent imputation network is based on RNNs, where we further model the temporal relations in the imputed data and exploit the uncertainties. While both GRU \cite{cho2014} (depicted in Fig. \ref{fig:grus_cells}a) and long-short term memory (LSTM) \cite{hochreiter1997} are feasible choices, inspired by the previous work of \mbox{GRU-D}\cite{che2018} depicted in Fig. \ref{fig:grus_cells}b, we employed a modified GRU cell by leveraging the uncertainty-aware GRU cell to consider further uncertainty and the temporal decaying factor, which is depicted in Fig. \ref{fig:grus_cells}c. 

Specifically, at each time step $t$, we produce the uncertainty decay factor $\vupsilon_t$ in Eq. (\ref{eq:unc_decay}) using a negative exponential rectifier to guarantee $\vupsilon_t \in (\mZero,\mOne]$ \cite{che2018,cao2018}.
\begin{equation}
    \label{eq:unc_decay}
    \vupsilon_t = \textrm{exp}\{-\textrm{max}(\mZero, \rmW_{\revise \bar{\rvu}} \bar{\rvu}_t + \rvb_{\revise \bar{\rvu}})\}
\end{equation}

\noindent We utilize such factors to emphasize the reliable estimates and give less attention to the uncertain ones. {\revise Hereafter, we denote $\rmW$ and $\rvb$ as the weights and biases of a fully-connected layer, respectively}. In particular, we first employ a fully-connected layer to $\bar{\rvx}_t$ and element-wise multiply with the uncertainty decay factor $\vupsilon_t$ as follows

\begin{equation}
    \label{eq:unc_decay_odot_x}
    \rvx^{{\revise \vupsilon}}_t = (\rmW_{\revise \vupsilon}\bar{\rvx}_t + \rvb_{\revise \vupsilon}) \odot \vupsilon_t.
\end{equation}

\noindent Note that we zeroed-out the diagonal of the parameter $\rmW_{\revise \vupsilon}$ to enforce the estimations based on other features. Thus, we obtain $\rvx^{{\revise \vupsilon}}_t$ as the \textit{feature-based correlated estimates} to the missing values.

In addition, we further consider the missing value estimates based on the temporal relations. For this purpose, we employ the time delay $\vdelta_t$ which is an essential element to capture temporal relations and missing patterns of the data \cite{che2018}. We exploit this information as the temporal decay factor $\vgamma_t \in (\mZero,\mOne]$ as follows
\begin{equation}
    \label{eq:time_decay}
    \vgamma_t = \textrm{exp}\{-\textrm{max}(\mZero, \rmW_{\revise \vgamma} \vdelta_t + \rvb_{\revise \vgamma})\}.
\end{equation}
Meanwhile, by employing the GRU, we obtain the hidden state $\rvh$ as the comprehensive information compiled from the preceding sequences. Thus, we take advantage of the temporal decay factor in governing the influence of past observations embedded into hidden states using the form of decayed hidden states as 
\begin{equation}
    \label{eq:decayed_hidden_states}
    \hat{\rvh}_{t-1} = \rvh_{t-1} \odot \vgamma_t.
\end{equation}
\noindent Thereby, given the previous hidden states $\hat{\rvh}_{t-1}$, we can estimate the current complete observation $\rvx^{{\revise \rvr}}_t$ through regression.    
\begin{equation}
    \label{eq:x_reg}
    \rvx^{{\revise \rvr}}_t = \rmW_{\revise \rvr}\hat{\rvh}_{t-1} + \rvb_{\revise \rvr}.
\end{equation}
In addition, we further make use of those estimates by applying another operation as

\begin{equation}
    \label{eq:x_temporal}
    \rvx^{{\revise \vtau}}_t = \rmW_{\revise \vtau}\rvx^{{\revise \rvr}}_t + \rvb_{\revise \vtau},
\end{equation}

\noindent again setting the diagonal parameter of $\rmW_{\revise \vtau}$ to be zeros to consider \textit{the feature-based estimation on top of the temporal relations from the previous hidden states}.

Hence, we have a pair of imputed values $\{\rvx^{{\revise \vupsilon}}_t, \rvx^{{\revise \vtau}}_t \}$, corresponding to missing value estimates obtained from the VAE by considering the uncertainties, and from the recurrent imputation network, respectively. We then merge this information jointly to get combined vector $\rvc_t$ comprising both estimates by simply employing a $1\times1$ convolution operation ($\ast$) as
\begin{equation}
    \label{eq:x_comb}
    \rvc_t = \rvx^{{\revise \vupsilon}}_t \ast \rvx^{{\revise \vtau}}_t
\end{equation}
Finally, we obtain the complete vector $\rvx^{{\revise \rvc}}_t$ by replacing the missing values with the combined estimates as 
\begin{equation}
    \label{eq:x_final}
    \rvx^{{\revise \rvc}}_t = \rvm_t \odot \tilde{\rvx}_t + (\mOne - \rvm_t) \odot \rvc_t.
\end{equation}
In addition, we concatenate the complete vector with the corresponding mask {\revise using $\circ$ operator}, and then feed it into the modified GRU cell to obtain the subsequent hidden states
\begin{equation}
    \label{eq:rnn}
    \rvh_t = \sigma(\rmW_{\revise \rvh}\hat{\rvh}_{t-1} + \rmV_{\revise \rvh}[\rvx^c_t \circ \rvm_t] + \rvb_{\revise \rvh})    
\end{equation}
{\revise with $\sigma$ denotes the activation function}. Lastly, to predict the in-hospital mortality as the clinical outcome, we utilize the last hidden state $\rvh_T$ to get the predicted label $\hat{y}$ such that
\begin{equation}
    \label{eq:output}
    \hat{y} = \sigma(\rmW_y\rvh_T + \rvb_y).
\end{equation}
Hereby, $\rmW_{\revise \{ \bar{\rvu}, \vupsilon, \vgamma, \rvr, \vtau, \rvh, y\}}$, $\rmV_{\revise \rvh}$, and $\rvb_{\revise \{\bar{\rvu}, \vupsilon, \vgamma, \rvr, \vtau, \rvh, y\}}$ are the learnable parameters in our recurrent imputation network.

\begin{algorithm}[t]
    \SetAlgoLined
    \SetKwInOut{Input}{input}\SetKwInOut{Output}{output}
    \Input{clinical time series data $\{\tilde{\rmX}, \rvy, \rmM, \mDelta\}$}
    \Output{imputed values $\rmX^{\revise \rvc}$; outcome prediction $\hat{y}$}
    $\nabla_{\vTheta} \leftarrow \mZero$ \\
    \While{not converge}{
        $q_\phi(\rmZ|\tilde{\rmX}) \leftarrow \mathcal{N}(\vmu_{\revise \rvz}, \textrm{diag}(\vsigma^2_{\revise \rvz}))$\\
        $\rmZ \leftarrow \vmu_{\revise \rvz} + \mSigma_{\revise \rvz} \odot \ \vepsilon_{\revise \rvz}, \quad \vepsilon_{\revise \rvz} \sim \mathcal{N}(\mZero,\mIdentity)$ \\
        $p_\theta(\tilde{\rmX}|\rmZ) \leftarrow \mathcal{N}(\vmu_{\revise \bar{\rvx}}, \textrm{diag}(\vsigma^2_{\revise \bar{\rvx}}))$ \\
        $\bar{\rmX} \leftarrow \rmM \odot \tilde{\rmX} + (\mOne - \rmM) \odot \vmu_{\revise \bar{\rvx}} $ \\
        $\bar{\rmU} \leftarrow (\mOne - \rmM) \odot \hat{\mSigma}_{\revise \bar{\rvx}}$\\
        $\rvh_0 \leftarrow \mZero$ \\
        \For{$t\leftarrow 1$ \KwTo $T$ }{
                // Eqs. (\ref{eq:unc_decay}) - (\ref{eq:rnn}) \\
                $\vupsilon_t \leftarrow \textrm{exp}\{-\textrm{max}(\mZero, \rmW_{\revise \bar{\rvu}} \bar{\rvu}_t + \rvb_{\revise \bar{\rvu}})\} $ \\
                $\rvx^{{\revise \vupsilon}}_t \leftarrow (\rmW_{\revise \vupsilon}\bar{\rvx}_t + \rvb_{\revise \vupsilon}) \odot \vupsilon_t$ \\
                $\vgamma_t \leftarrow \textrm{exp}\{-\textrm{max}(\mZero, \rmW_{\revise \vgamma} \vdelta_t + \rvb_{\revise \vgamma})\}$ \\                
                $\hat{\rvh}_{t-1} \leftarrow \rvh_{t-1} \odot {\revise \vgamma}_t$ \\
                $\rvx^{{\revise \rvr}}_t \leftarrow \rmW_{\revise \rvr}\hat{\rvh}_{t-1} + \rvb_{\revise \rvr}$ \\
                $\rvx^{{\revise \vtau}}_t \leftarrow \rmW_{\revise \vtau}\rvx^{{\revise \rvr}}_t + \rvb_{\revise \vtau}$ \\
                $\rvc_t \leftarrow \rvx^{\revise \vupsilon}_t \ast \rvx^{{\revise \vtau}}_t$ \\
                $\rvx^{{\revise \rvc}}_t \leftarrow \rvm_t \odot \bar{\rvx}_t + (\mOne - \rvm_t) \odot \rvc_t$ \\ 
                $\rvh_{t} \leftarrow \textrm{GRU}([\rvx^{{\revise \rvc}}_t \circ \rvm_t],\hat{\rvh}_{t-1})$
        }
    $\hat{y} \leftarrow \sigma(\rmW_y\rvh_T + \rvb_y)$ \\
    $\gL_{\textrm{total}} \leftarrow $ Eq. (\ref{eq:loss_total}) \\
    $\nabla_{\vTheta} \leftarrow \frac{\partial}{\partial{\vTheta}} \gL_{\textrm{total}}$ \\
    ${\vTheta} \leftarrow \textrm{optimize}({\vTheta},\nabla_{\vTheta})$ \\
    }
    \caption{Algorithm of our proposed model}
    \label{algo:model}
\end{algorithm}

\subsection{Learning}
We describe the composite loss function, comprising the imputation and prediction loss function to tune all model parameters jointly, which are $\vTheta = \{\theta, \phi, \rmW_{\revise \{\bar{\rvu}, \vupsilon, \vgamma, \rvr, \vtau, \rvh, y\}}, \rmV_{\revise \rvh}, \rvb_{\revise \{\bar{\rvu}, \vupsilon, \vgamma, \rvr, \vtau, \rvh, y\}}\}$. Such loss function accommodates the VAEs and the recurrent imputation network as well. By means of VAEs, we define the loss function $\gL_{vae}$ to maximize the variational evidence lower bound (ELBO) that comprises the reconstruction loss term and the {\revise Kullback-Leibler divergence \cite{kullbackleibler1951}}. We add $\ell_1$-regularization to introduce sparsity into the network with $\lambda_1$ as the hyperparameter. Moreover, for each time step, we measure the difference between the observed data and the combined imputation estimates by the mean absolute error (MAE) as $\gL_{reg}$.
\begin{equation*}
    \label{eq:loss_vae}
    \gL_{vae} =  \sum^N_{n=1} \sum^T_{t=1} \mathbb{E}_{q_\phi(\rvz|\tilde{\rvx}^{(n)}_t)}[\log p_\theta(\tilde{\rvx}^{(n)}_t|\rvz)]
\end{equation*}
\begin{equation}
    -  \KLDiv{q_\phi(\rvz|\tilde{\rvx}^{(n)}_t)}{p_\theta(\rvz)} + \lambda_1 \lVert(\theta;\phi)\lVert_1
\end{equation}
\begin{equation}
    \label{eq:loss_reg}
    \gL_{reg} = \sum^N_{n=1}{\sum^T_{t=1}{ \gL_{MAE}(\tilde{\rvx}^{(n)}_t \odot \rvm^{(n)}_t, \rvc^{(n)}_t \odot \rvm^{(n)}_t)}} 
\end{equation}
Furthermore, we define the binary cross-entropy loss function $\gL_{pred}$ to evaluate the prediction of in-hospital mortality.
Thus, we define the overall composite loss function $\gL_{total}$ as
\begin{equation}
    \label{eq:loss_total}
    \gL_{total} = \alpha \ \gL_{vae} + \beta \ \gL_{reg} + \gL_{pred}, 
\end{equation}
where $\alpha$ and $\beta$  are the hyperparameters to represent the ratio between the $\gL_{vae}$ and $\gL_{reg}$, respectively. 

Note that our proposed model is also applicable to consider the bidirectional dynamics. Such a scenario can be carried out by having the forward and backward direction of the data fed into the recurrent imputation network. By doing so, we make our proposed model a fair comparison to M-RNN \cite{yoon2017}, BRITS-I {\revise \cite{cao2018}}, and BRITS \cite{cao2018} which adopted such strategy to achieve better estimates of the missing values and the prediction outcomes. {\revise In bidirectional cases, we employ an additional consistency loss function \cite{cao2018} to $\gL_{total}$ in order to impose a consistency estimates for each time step in both directions as}
\begin{equation}
    \label{eq:loss_consistency}
    \gL_{cons} = \sum^N_{n=1}{\sum^T_{t=1}{ \gL_{MAE}(\rvx^{{\revise \rvc},(n)}_t, \rvx^{{\revise \rvc'},(n)}_t)}} ,
\end{equation}
\noindent with $\rvx^{{\revise \rvc},(n)}_t$ and $\rvx^{{\revise \rvc'},(n)}_t$ denoting the estimates from the forward and backward direction, respectively.
Another hyperparameter $\xi$ could be introduced for this consistency loss to optimize the model. Lastly, we use stochastic gradient descent in an end-to-end manner to optimize the model parameters during the training. We summarize the overall training steps of our proposed framework in Algorithm \ref{algo:model}.


\section{Experiments} 
\label{section:experiments}
\subsection{Dataset and Implementation Setup}
\label{section:experiments_dataset}
{\bf PhysioNet 2012 Challenge} \cite{goldberger2000,ikaro2012} consists of 35 irregularly sampled clinical variables (\eg, \ heart and respiration rate, blood pressure, \etc) from 4,000 patients during their first 48 hours of medical care in the ICU. Note that we ignore the demographic information and categorical data types from this dataset. Hereby, we exploit only the clinical time series data. From those samples, we excluded three patients with no observations at all. We sampled the observations hourly, using the time window as the timestamps, and took the average of values in cases of multiple measurements within this time window. It resulted in sparse EHR data with an average missing rate of 80.51\%. Our aim is to predict the in-hospital mortality of patients, with 554 positive mortality labels (13.86\%). As for the implementation setup of the PhysioNet dataset, we employed three layers of feedforward networks for the inference network of VAEs with hidden units of $\{64,24,10\}$, with $10$ denoting the dimension of latent representation. The generative network has equal numbers of hidden units with those of the inference network but in the reverse order. We employed hyperbolic tangent (tanh) as the non-linear activation function for each hidden layer. Prior to those activation functions, we also applied batch normalization and a dropout rate of $\{0.1,0.3\}$ for classification and imputation tasks, respectively. We employed modified GRU for recurrent imputation network with $64$ hidden units. 

{\bf 
The Medical Information Mart for Intensive Care (MIMIC-III)} \cite{johnson2016} dataset consists of 53,432 ICU stays of adult patients in the Beth Israel Deaconess Medical Center in the period of 2001 --  2012 \cite{johnson2016}. We selected 99 variables from several source tables, such as laboratory tests, inputs to patients, outputs from patients, and drug prescriptions tables, resulting in a cohort of 13,998 patients with 1,181 positive in-hospital mortality labels (8.44\%) among them. Moreover, from those irregular measurements, we further sampled the data into two hourly samples for the first 48 hours of their medical care, leading to an average missing rate of 93.92\%. As in the case of PhysioNet, we took the average value if there existed multiple measurements. We referred to \cite{che2018,purushotham2018} in pre-processing the MIMIC-III cohort. We employed three layers of feedforward networks for the inference network of VAEs with hidden units of $\{128,32,16\}$ and an equal number of hidden units in reverse for the generative network. Likewise, for each hidden layer, we used batch normalization, drop out rate of $0.3$ for both of classification and imputation tasks, and tanh activation function. Furthermore, $64$ hidden units were employed for the recurrent imputation network. 

We trained the proposed model on both datasets using an Adam optimizer with $100$ epochs and $64$ mini-batches. For imputation task, we fixed the learning rates of $0.0005$ and $0.0003$ for PhysioNet and MIMIC-III, respectively, while $0.005$ for both datasets on the classification task. We set $\lambda_{1}$ and weight decay equally with $1e-5$. {\revise As for} bidirectional models, we set $\xi=0.1$ as the hyperparameter of the consistency loss.


\subsection{Tasks}
\label{section:experiments_task}
In this work, we validated the performance of our proposed models from two perspectives: (1) the in-hospital mortality prediction (classification), and (2) missing value imputation.

\subsubsection{Classification Task}
Our primary goal for this work is to predict the in-hospital mortality as the binary classification task. For this purpose, we reported the test result on the in-hospital mortality prediction task from the 5-fold cross-validation in terms of the average of the  Area  Under the  Curve  (AUC) ROC. Additionally, to measure the robustness of the models in dealing with the imbalance data portrayed in both datasets, we also reported the Area Under the Precision-Recall Curve (AUPRC). We randomly removed samples of the training data with $\{5\%, 10\%\}$ scenarios and left the validation and test set untouched to be reported as the results. 

\subsubsection{Imputation Task}
For the secondary task, we additionally evaluated imputation performance on the missing values. As for this task, we randomly removed samples with settings of $\{5\%, 10\%\}$ of observed data from all training, validation, and test set as the ground truth. Then we reported the test result from the 5-fold cross-validation by measuring the MAE. In addition, we also measured the other most-used imputation similarity metric, {\revise namely mean relative error (MRE) and mean squared error (MSE)}. Given $\rvx^{(i)}$ and $\rvx^{{\revise \rvc},(i)}$ as the ground truth and imputation estimates of $i$-th item, respectively, and $N_{GT}$ ground truth items in total, {\revise we defined MAE, MRE, and MSE as}
{\revise
\begin{equation}
    \textrm{MAE} = \frac{\sum_{i=1}^{N_{GT}} |\rvx^{{\revise \rvc},(i)} - \rvx^{(i)}|}{N_{GT}},
\end{equation}
\begin{equation}
    \textrm{MRE} = \frac{\sum_{i=1}^{N_{GT}} |\rvx^{{\revise \rvc},(i)} - \rvx^{(i)}|}{\sum_{i=1}^{N_{GT}} |\rvx^{(i)}|},
\end{equation}
\begin{equation}
    \textrm{MSE} = \frac{\sum_{i=1}^{N_{GT}} (\rvx^{{\revise \rvc},(i)} - \rvx^{(i)})^2}{N_{GT}}.
\end{equation}
}
\begin{figure*}[t]
\centering
\includegraphics[width=0.95\textwidth]{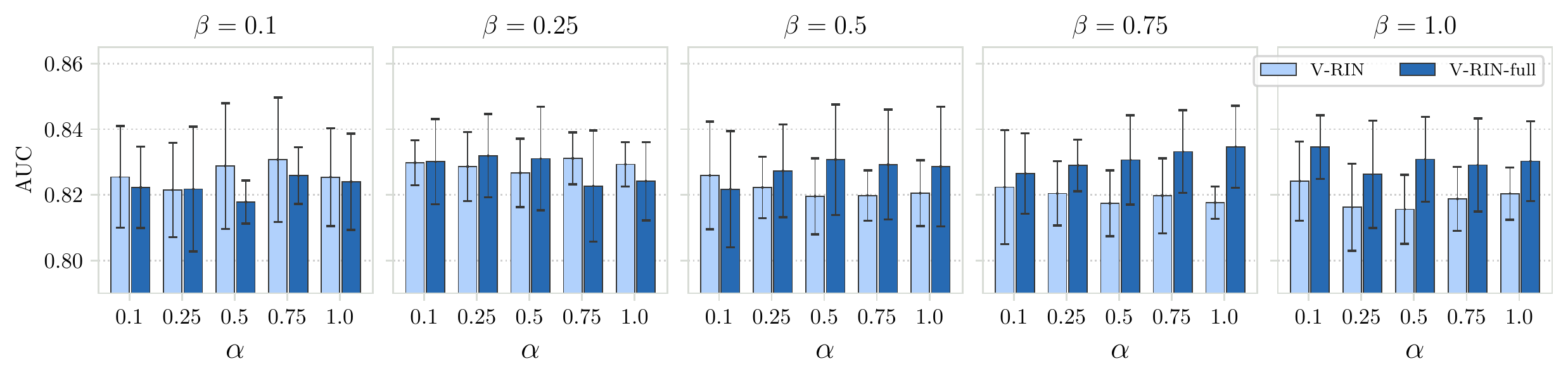}
\caption{Ablation studies on the impact of hyperparameter pair $\{\alpha, \beta\}$ for the classification task on PhysioNet dataset.}
\label{fig:ablation_study_physionet}
\end{figure*}

\begin{table*}[t]
\caption{\revise Ablation study results on both PhysioNet and MIMIC-III datasets using 10\% removal scenario. *) For the VAE+RNN \cite{jun2019}, we obtained the missing value estimates solely from the VAE in reconstructing values}
\label{tab:ablation_studies}
\centering
\begin{tabular}{cccllll}
\toprule
\multicolumn{1}{c}{\textbf{Dataset}} &
\multicolumn{1}{c}{\textbf{Task}} &
\multicolumn{1}{c}{\textbf{Metric}} & \multicolumn{1}{c}{\textbf{VRNN \cite{chung2015}}} &
\multicolumn{1}{c}{\textbf{VAE+RNN \cite{jun2019}}} &
\multicolumn{1}{c}{\textbf{V-RIN (Ours)}} &
\multicolumn{1}{c}{\textbf{V-RIN-full (Ours)}} \\
\midrule
\multirow{4}{*}{PhysioNet} & \multirow{2}{*}{\textit{Classification}} 
& \multicolumn{1}{l}{AUC} 
& \multicolumn{1}{c}{$0.7964 \pm 0.0117$} 
&  \multicolumn{1}{c}{$0.8098 \pm 0.0154$} 
& \multicolumn{1}{c}{$0.8311 \pm 0.0079$} 
&  \multicolumn{1}{c}{${\bf 0.8347 \pm 0.0125}$}\\
& & \multicolumn{1}{l}{AUPRC} 
& \multicolumn{1}{c}{$0.4082 \pm 0.0369$}  
& \multicolumn{1}{c}{$0.4311 \pm 0.0409$} 
& \multicolumn{1}{c}{$0.4781 \pm 0.0529$} 
& \multicolumn{1}{c}{${\bf 0.4792 \pm 0.0277}$} \\  \cmidrule{2-7}
& \multirow{2}{*}{\textit{Imputation}} 
& \multicolumn{1}{l}{MAE} 
& \multicolumn{1}{c}{$0.6948 \pm 0.0171$} 
& \multicolumn{1}{c}{$0.6493 \pm 0.0226^*$} 
& \multicolumn{1}{c}{$0.2938 \pm 0.0080$} 
& \multicolumn{1}{c}{${\bf 0.2898 \pm 0.0081}$} \\ 
& & \multicolumn{1}{l}{MRE} 
& \multicolumn{1}{c}{$0.8790 \pm 0.0186$} 
& \multicolumn{1}{c}{$0.9126 \pm 0.0203^*$} 
& \multicolumn{1}{c}{$0.4130 \pm 0.0049$} 
& \multicolumn{1}{c}{${\bf 0.40727 \pm 0.0055}$} \\ 
& & \multicolumn{1}{l}{\revise MSE} 
& \multicolumn{1}{c}{\revise $1.0910 \pm 0.0200$} 
& \multicolumn{1}{c}{\revise $0.8538 \pm 0.0906^*$} 
& \multicolumn{1}{c}{\revise $0.3889 \pm 0.0579$} 
& \multicolumn{1}{c}{\revise ${\bf 0.3807 \pm 0.0546}$} \\ 

\midrule
\multirow{4}{*}{MIMIC-III} & \multirow{2}{*}{\textit{Classification}} 
& \multicolumn{1}{l}{AUC} &
\multicolumn{1}{c}{$0.7925 \pm 0.0126$} & 
\multicolumn{1}{c}{$0.8147 \pm 0.0140$} & 
\multicolumn{1}{c}{$0.8634 \pm 0.0137$} & 
\multicolumn{1}{c}{${\bf 0.8650 \pm 0.0091}$} \\
& & \multicolumn{1}{l}{AUPRC} 
& \multicolumn{1}{c}{$0.3087 \pm 0.0291$} 
& \multicolumn{1}{c}{$0.3482 \pm 0.0159$} 
& \multicolumn{1}{c}{$0.4099 \pm 0.0163$} 
& \multicolumn{1}{c}{${\bf 0.4144 \pm 0.0117}$} \\ \cmidrule{2-7}
& \multirow{2}{*}{\textit{Imputation}}
& \multicolumn{1}{l}{MAE} 
& \multicolumn{1}{c}{$0.7042 \pm 0.0048$} 
& \multicolumn{1}{c}{$0.6268 \pm 0.0063^*$} 
& \multicolumn{1}{c}{$0.3322 \pm 0.0072$} 
& \multicolumn{1}{c}{${\bf 0.3279 \pm 0.0055}$} \\ 
& & \multicolumn{1}{l}{MRE} 
& \multicolumn{1}{c}{$0.9922 \pm 0.0026$} 
& \multicolumn{1}{c}{$0.9947 \pm 0.0011^*$} 
& \multicolumn{1}{c}{$0.5271 \pm 0.0080$} 
& \multicolumn{1}{c}{${\bf 0.5204 \pm 0.0065}$} \\ 
& & \multicolumn{1}{l}{\revise MSE} 
& \multicolumn{1}{c}{\revise $1.0661 \pm 0.0208$} 
& \multicolumn{1}{c}{\revise $1.0523 \pm 0.1110^*$} 
& \multicolumn{1}{c}{\revise $0.6172 \pm 0.1174$} 
& \multicolumn{1}{c}{\revise ${\bf 0.5865 \pm 0.1038}$} \\ 
\bottomrule
\end{tabular}
\end{table*}

\subsection{Comparative Models}
\label{section:experiments_compmodels}
We compared the performance of our proposed model in carrying the aforementioned tasks with the closely-related competing state-of-the-art models in the literature by grouping them into unidirectional and bidirectional models. 
\subsubsection{Unidirectional Models}
\begin{itemize}
    \item {{\bf GRU-D} \cite{che2018} estimates the missing values by utilizing the informative missing value patterns in the form of the masking and time decay factor using modified GRU cells.}
    \item {{\bf RITS-I} \cite{cao2018} utilizes unidirectional dynamics that rely solely on temporal relations through regression.}
    \item {{\bf RITS} \cite{cao2018} is devised based on RITS-I by further taking into account the feature correlations function. Furthermore, it utilizes the temporal decay as the factor to weigh between both features and temporal-based estimates.}
    \item {{\bf V-RIN} (Ours) is based on our proposed model except that we ignored the uncertainty decay factor. Specifically, we excluded Eq. (\ref{eq:unc_decay}) and omitted the element-wise multiplication operation of the $\vupsilon_t$ in Eq. (\ref{eq:unc_decay_odot_x}).}
    \item {{\bf V-RIN-full} (Ours) executed all operations in the proposed model including feature-based correlations, temporal relations, and the uncertainty decay utilization. }
\end{itemize}
\subsubsection{Bidirectional Models}
\begin{itemize}
    \item {{\bf M-RNN} \cite{yoon2017} exploits the multi-directional RNNs which execute both interpretation and imputation.} 
    \item{{\bf BRITS-I} \cite{cao2018} is based on the RITS-I by extending it to be able to handle bidirectional dynamics in estimating the missing values.}
    \item {{\bf BRITS} \cite{cao2018} takes the bidirectional dynamics of RITS in handling the sparsity in the data. Both \mbox{BRITS-I} and BRITS additionally employ consistency loss of forward and backward directions in their attempt to estimate the missing values more precisely.}
    \item {We extended our proposed model of  {\bf V-RIN} and {\bf V-RIN-full} into the bidirectional models by means of the recurrent imputation networks. Similar to BRITS-I and BRITS, we further computed consistency loss in Eq. (\ref{eq:loss_consistency}). }
\end{itemize}

\begin{table*}[t]
\caption{\revise Performance of in-hospital mortality prediction task on both PhysioNet and MIMIC-III Datasets}
\label{tab:classifcation_task}
\centering
\begin{tabular}{ccccccc}
\toprule
\multirow{2}{*}{\textbf{Dataset}} & \multicolumn{2}{c}{\multirow{2}{*}{\textbf{Models}}} & \multicolumn{2}{c}{\textbf{10\%}} & \multicolumn{2}{c}{\textbf{5\%}}\\
\cmidrule{4-7} 
& & & \textbf{AUC} & \textbf{AUPRC} & \textbf{AUC} & \textbf{AUPRC} \\
\midrule
\multirow{10}{*}{PhysioNet} &
\multirow{5}{*}{\rotatebox[origin=c]{90}{\textit{Unidirectional}}}&
\multicolumn{1}{l}{GRU-D \cite{che2018}} & 
\multicolumn{1}{c}{$0.8096 \pm 0.0308$} & 
\multicolumn{1}{c}{$0.4645 \pm 0.0464$} & 
\multicolumn{1}{c}{$0.8096 \pm 0.0308$} & 
\multicolumn{1}{c}{$0.4645 \pm 0.0464$} 
\\ 
&& \multicolumn{1}{l}{RITS-I \cite{cao2018}} & 
\multicolumn{1}{c}{$0.8160 \pm 0.0108$} & 
\multicolumn{1}{c}{$0.4477 \pm 0.0395$} & 
\multicolumn{1}{c}{$0.8179 \pm 0.0110$} & 
\multicolumn{1}{c}{$0.4385	\pm 0.0222$}
\\
&& \multicolumn{1}{l}{RITS \cite{cao2018}} & 
\multicolumn{1}{c}{$0.8159 \pm 0.0153$} & 
\multicolumn{1}{c}{$0.4484 \pm 0.0361$} & 
\multicolumn{1}{c}{$0.8109 \pm 0.0159$} & 
\multicolumn{1}{c}{$0.4568	\pm 0.0394$}
\\
&& \multicolumn{1}{l}{\bf V-RIN (Ours)} &
\multicolumn{1}{c}{$0.8311 \pm 0.0079$} &
\multicolumn{1}{c}{$0.4781 \pm 0.0529$} & 
\multicolumn{1}{c}{$0.8292 \pm 0.0127$} &
\multicolumn{1}{c}{${\bf 0.4773 \pm 0.0428}$}
\\
&& \multicolumn{1}{l}{\bf V-RIN-full (Ours)} &
\multicolumn{1}{c}{${\bf 0.8347 \pm 0.0125}$} &
\multicolumn{1}{c}{${\bf 0.4792 \pm 0.0277}$} &
\multicolumn{1}{c}{${\bf 0.8348 \pm 0.0116}$} &
\multicolumn{1}{c}{${\bf 0.4773 \pm 0.0374}$}
\\ 
\cmidrule{2-7}
&\multirow{5}{*}{\rotatebox[origin=c]{90}{\textit{Bidirectional}}}&
\multicolumn{1}{l}{M-RNN \cite{yoon2017}} & 
\multicolumn{1}{c}{$0.7885 \pm 0.0236$} & 
\multicolumn{1}{c}{$0.4096 \pm 0.0296$} & 
\multicolumn{1}{c}{$0.7817 \pm 0.0230$} & 
\multicolumn{1}{c}{$0.3839 \pm 0.0426$}  
\\
&& \multicolumn{1}{l}{BRITS-I \cite{cao2018}} & 
\multicolumn{1}{c}{$0.8253 \pm 0.0215$} & 
\multicolumn{1}{c}{$0.4521 \pm 0.0347$} &  
\multicolumn{1}{c}{$0.8168 \pm 0.0175$} & 
\multicolumn{1}{c}{$0.4508	\pm 0.0327$}
\\
&& \multicolumn{1}{l}{BRITS \cite{cao2018}} & 
\multicolumn{1}{c}{$0.8225 \pm 0.0134$} & 
\multicolumn{1}{c}{$0.4580 \pm 0.0284$} & 
\multicolumn{1}{c}{$0.8242	 \pm 0.0041$} & 
\multicolumn{1}{c}{$0.4600 \pm 0.0420$} 
\\ 
&& \multicolumn{1}{l}{\bf V-RIN (Ours)} &
\multicolumn{1}{c}{$0.8393 \pm 0.0177$} &
\multicolumn{1}{c}{$0.4760 \pm 0.0496$} &
\multicolumn{1}{c}{$0.8377 \pm 0.0127$} &
\multicolumn{1}{c}{${\bf 0.4886 \pm 0.0402}$} 
\\
&& \multicolumn{1}{l}{\bf V-RIN-full (Ours)} &
\multicolumn{1}{c}{${\bf 0.8401 \pm 0.0100}$} &
\multicolumn{1}{c}{${\bf 0.4924 \pm 0.0363}$} &
\multicolumn{1}{c}{${\bf 0.8422 \pm 0.0070}$} &
\multicolumn{1}{c}{$0.4871 \pm 0.0217$} 
\\
\midrule
\multirow{10}{*}{MIMIC-III} &
\multirow{5}{*}{\rotatebox[origin=c]{90}{\textit{Unidirectional}}}&
\multicolumn{1}{l}{GRU-D \cite{che2018}} & 
\multicolumn{1}{c}{$0.8536 \pm 0.0138$} & 
\multicolumn{1}{c}{$0.3809	\pm 0.0198$} &  
\multicolumn{1}{c}{$0.8536 \pm 0.0138$} & 
\multicolumn{1}{c}{$0.3809 \pm 0.0198$}
\\ 
&& \multicolumn{1}{l}{RITS-I \cite{cao2018}} & 
\multicolumn{1}{c}{$0.8533 \pm 0.0127$} & 
\multicolumn{1}{c}{$0.3687 \pm 0.0119$} &  
\multicolumn{1}{c}{$0.8543 \pm 0.0067$} &
\multicolumn{1}{c}{$0.3707 \pm 0.0119$}
\\
&& \multicolumn{1}{l}{RITS \cite{cao2018}} & 
\multicolumn{1}{c}{$0.8601 \pm 0.0044$} & 
\multicolumn{1}{c}{$0.3922 \pm 0.0048$} &  
\multicolumn{1}{c}{$0.8547 \pm 0.0091$} & 
\multicolumn{1}{c}{$0.3739 \pm 0.0289$} 
\\
&& \multicolumn{1}{l}{\bf V-RIN (Ours)} &
\multicolumn{1}{c}{$0.8634 \pm 0.0137$} &
\multicolumn{1}{c}{$0.4099 \pm 0.0163$} &
\multicolumn{1}{c}{${\bf 0.8620 \pm 0.0112}$} &
\multicolumn{1}{c}{${\bf 0.3951 \pm 0.0185}$}
\\
&& \multicolumn{1}{l}{\bf V-RIN-full (Ours)} &
\multicolumn{1}{c}{${\bf 0.8650 \pm 0.0091}$} &
\multicolumn{1}{c}{${\bf 0.4144 \pm 0.0117}$} &
\multicolumn{1}{c}{$0.8616 \pm 0.0101$} &
\multicolumn{1}{c}{$0.3924 \pm 0.0220$}
\\
\cmidrule{2-7}
&\multirow{5}{*}{\rotatebox[origin=c]{90}{\textit{Bidirectional}}}&
\multicolumn{1}{l}{M-RNN \cite{yoon2017}} & 
\multicolumn{1}{c}{$0.8243 \pm 0.0065$} & 
\multicolumn{1}{c}{$0.3221 \pm 0.0235$} & 
\multicolumn{1}{c}{$0.8227 \pm 0.0101$} & 
\multicolumn{1}{c}{$0.3176 \pm 0.0162$} 
\\ 
&& \multicolumn{1}{l}{BRITS-I \cite{cao2018}} & 
\multicolumn{1}{c}{$0.8625 \pm 0.0093$} & 
\multicolumn{1}{c}{$0.4283 \pm 0.0359$} & 
\multicolumn{1}{c}{$0.8621 \pm 0.0089$} & 
\multicolumn{1}{c}{$0.4065 \pm 0.0159$} 
\\
&& \multicolumn{1}{l}{BRITS \cite{cao2018}} & 
\multicolumn{1}{c}{$0.8649 \pm 0.0110$} & 
\multicolumn{1}{c}{$0.4129 \pm 0.0324$} & 
\multicolumn{1}{c}{$0.8638 \pm 0.0141$} & 
\multicolumn{1}{c}{$0.4144 \pm 0.0250$} 
\\
&& \multicolumn{1}{l}{\bf V-RIN (Ours)} &
\multicolumn{1}{c}{${\bf 0.8695 \pm 0.0139}$} &
\multicolumn{1}{c}{${\bf 0.4333 \pm 0.0288}$} &
\multicolumn{1}{c}{$0.8694 \pm 0.0121$} &
\multicolumn{1}{c}{$0.4158 \pm 0.0204$}  
\\
&& \multicolumn{1}{l}{\bf V-RIN-full (Ours)} &
\multicolumn{1}{c}{$0.8691 \pm 0.0126$} &
\multicolumn{1}{c}{$0.4270 \pm 0.0255$} &
\multicolumn{1}{c}{${\bf 0.8705 \pm 0.0148}$} &
\multicolumn{1}{c}{${\bf 0.4199 \pm 0.0160}$}
\\
\bottomrule
\end{tabular}
\end{table*}

\subsection{Experimental Results : Ablation Studies}
\label{section:experiments_ablation}
As part of the ablation studies, we report the performance of the unidirectional model of the V-RIN and V-RIN-full models on the in-hospital mortality classification task. Firstly, we investigated the effect of the pair $\{\alpha, \beta\}$ as the hyperparameter in Eq. (\ref{eq:loss_total}). We reflected these parameters as a ratio to weigh the imputation by the VAEs and the recurrent imputation network to achieve the optimal performance on estimating the missing values and classifying the clinical outcomes. For each parameter, we defined set of values in the range of $[0.1, 1.0]$. 

For PhysioNet, as illustrated in Fig. \ref{fig:ablation_study_physionet}, in general, we observed that in almost all the combination settings, V-RIN-full achieved higher performance than V-RIN. We interpreted these findings that   introducing the uncertainty helps the model in estimating the missing values, leading to better classifying the outcome. Both models were able to achieve high performance in terms of their average AUC scores of $0.8311 \pm 0.0079$ for V-RIN and $0.8347 \pm 0.0125$ for V-RIN-full. These AUC results were obtained with settings of $\{\alpha=0.75,\ \beta=0.25\}$ and $\{\alpha=1.0,\ \beta=0.75\}$, respectively. For these results, we observed that the model favored the emphasis on the feature correlations over the temporal relations to obtain its best performance. For the case of V-RIN, once we tried to increase the $\beta>0.25$, we observed that the classification performance was degraded to some degree. In contrast, the performance of V-RIN-full was considerably better when we increased the $\beta$ parameter. We also carried out similar ablation studies on the MIMIC-III dataset, which is reported in the supplementary material. To summarize, for both datasets, we argue that both features and temporal relations are essential in estimating the missing values with some latent proportion. 
 
Table \ref{tab:ablation_studies} presents the comparison of our model with closely related models on both classification and imputation tasks, such as VRNN \cite{chung2015}, which integrates VAEs for each time step of RNNs, and VAE+RNN \cite{jun2019}, which employs VAEs followed by RNNs without incorporating the uncertainty. For the case of VRNN on PhysioNet and MIMIC-III, we observed that the classification performance was the lowest among reported models in terms of both its AUC and AUPRC. In addition, their imputation performance was {\revise low} in comparison to both our proposed models. As in the case of VAE+RNN, in comparison to VRNN, we noticed a considerable improvement in the performance on PhysioNet and MIMIC-III, especially in terms of AUPRC. VAE+RNN is, in fact, the model closely related to ours that it executes the imputation process by first exploiting the feature correlations followed by temporal dynamics in exact order. However, \cite{jun2019} employed the vanilla RNNs instead of recurrent imputation network, which is a novel extension in this study. By introducing the temporal decay in V-RIN, the model was better able to learn the temporal dynamics effectively, resulting in better AUC and AUPRC, as well as better imputation results in terms of {\revise MAE, MRE and MSE} on both datasets, by a large margin. Finally, once we introduced the uncertainty which is incorporated in the recurrent imputation network of V-RIN-full, we observed a significant enhancement of {\revise overall performance metrics on both tasks and datasets}. Thus, we conclude that the utilization of both temporal decay and the uncertainties are beneficial in both imputation and classification tasks.

\begin{table*}[t]
\caption{\revise Performance of imputation task on both PhysioNet and MIMIC-III Datasets}
\label{tab:imputation_task}
\centering
\begin{tabular}{cccccccc}
\toprule
\multicolumn{2}{c}{\multirow{2}{*}{\textbf{Models}}} & \multicolumn{3}{c}{\textbf{10\%}} & \multicolumn{3}{c}{\textbf{5\%}}\\
\cmidrule{3-8} 
& & \textbf{MAE} & \textbf{MRE} & \textbf{\revise MSE} & \textbf{MAE} & \textbf{MRE} & \textbf{\revise MSE} \\
\midrule
\multicolumn{8}{c}{PhysioNet} \\
\midrule
\multirow{5}{*}{\rotatebox[origin=c]{90}{\textit{Unidirectional}}}& \multicolumn{1}{l}{GRU-D \cite{che2018}} & 
\multicolumn{1}{c}{$0.661 \pm 0.016$} & 
\multicolumn{1}{c}{$0.930 \pm 0.014$} &  
\multicolumn{1}{c}{$0.900 \pm 0.107$} &  
\multicolumn{1}{c}{$0.662 \pm 0.013$} & 
\multicolumn{1}{c}{$0.929 \pm 0.015$} &
\multicolumn{1}{c}{$0.906 \pm 0.105$}   
\\ 
& \multicolumn{1}{l}{RITS-I \cite{cao2018}} & 
\multicolumn{1}{c}{$0.407 \pm 0.009$} & 
\multicolumn{1}{c}{$0.572 \pm 0.004$} &
\multicolumn{1}{c}{$0.476 \pm 0.060$} &  
\multicolumn{1}{c}{$0.415 \pm 0.010$} & 
\multicolumn{1}{c}{$0.582 \pm 0.009$} &
\multicolumn{1}{c}{$0.507 \pm 0.087$} 
\\
& \multicolumn{1}{l}{RITS \cite{cao2018}} & 
\multicolumn{1}{c}{$0.307 \pm 0.008$} & 
\multicolumn{1}{c}{$0.431 \pm 0.005$} & 
\multicolumn{1}{c}{$0.390 \pm 0.057$} & 
\multicolumn{1}{c}{$0.327 \pm 0.008$} & 
\multicolumn{1}{c}{$0.459 \pm 0.007$}&
\multicolumn{1}{c}{$0.431 \pm 0.074$}
\\
& \multicolumn{1}{l}{\bf V-RIN (Ours)} &
\multicolumn{1}{c}{$0.293 \pm 0.008$} &
\multicolumn{1}{c}{$0.412 \pm 0.004$} &
\multicolumn{1}{c}{$0.388 \pm 0.057$} & 
\multicolumn{1}{c}{$0.320 \pm 0.008$} &
\multicolumn{1}{c}{$0.449 \pm 0.007$} &
\multicolumn{1}{c}{$0.426 \pm 0.077$}
\\
& \multicolumn{1}{l}{\bf V-RIN-full (Ours)} &
\multicolumn{1}{c}{${\bf 0.289 \pm 0.008}$} &
\multicolumn{1}{c}{${\bf 0.407 \pm 0.005}$} &
\multicolumn{1}{c}{${\bf 0.380 \pm 0.054}$} & 
\multicolumn{1}{c}{${\bf 0.315 \pm 0.008}$} &
\multicolumn{1}{c}{${\bf 0.442 \pm 0.009}$} &
\multicolumn{1}{c}{${\bf 0.422 \pm 0.075}$}
\\ 
\midrule
\multirow{5}{*}{\rotatebox[origin=c]{90}{\textit{Bidirectional}}}&
\multicolumn{1}{l}{M-RNN \cite{yoon2017}} & 
\multicolumn{1}{c}{$0.397 \pm 0.013$} & 
\multicolumn{1}{c}{$0.558 \pm 0.009$} &  
\multicolumn{1}{c}{$0.439 \pm 0.066$} & 
\multicolumn{1}{c}{$0.411 \pm 0.011$} & 
\multicolumn{1}{c}{$0.577 \pm 0.010$} &
\multicolumn{1}{c}{$0.481 \pm 0.088$}
\\ 
& \multicolumn{1}{l}{BRITS-I \cite{cao2018}} & 
\multicolumn{1}{c}{$0.371 \pm 0.011$} & 
\multicolumn{1}{c}{$0.521 \pm 0.008$} &  
\multicolumn{1}{c}{$0.416 \pm 0.064$} & 
\multicolumn{1}{c}{$0.380 \pm 0.010$} & 
\multicolumn{1}{c}{$0.534 \pm 0.010$} &
\multicolumn{1}{c}{$0.448 \pm 0.085$}
\\
& \multicolumn{1}{l}{BRITS \cite{cao2018}} & 
\multicolumn{1}{c}{$0.280 \pm 0.009$} & 
\multicolumn{1}{c}{$0.393 \pm 0.006$} & 
\multicolumn{1}{c}{$0.351 \pm 0.058$} & 
\multicolumn{1}{c}{$0.298 \pm 0.008$} & 
\multicolumn{1}{c}{$0.418 \pm 0.007$} &
\multicolumn{1}{c}{$0.390 \pm 0.076$}
\\
& \multicolumn{1}{l}{\bf V-RIN (Ours)} &
\multicolumn{1}{c}{$0.268 \pm 0.008$} &
\multicolumn{1}{c}{$0.375 \pm 0.006$} &
\multicolumn{1}{c}{$0.348 \pm 0.059$} & 
\multicolumn{1}{c}{$0.292 \pm 0.008$} &
\multicolumn{1}{c}{$0.411 \pm 0.008$} &
\multicolumn{1}{c}{$0.387 \pm 0.075$}
\\
& \multicolumn{1}{l}{\bf V-RIN-full (Ours)} &
\multicolumn{1}{c}{${\bf 0.264 \pm 0.008}$} &
\multicolumn{1}{c}{${\bf 0.371 \pm 0.006}$} &
\multicolumn{1}{c}{${\bf 0.343 \pm 0.059}$} & 
\multicolumn{1}{c}{${\bf 0.289 \pm 0.008}$} &
\multicolumn{1}{c}{${\bf 0.406 \pm 0.008}$} &
\multicolumn{1}{c}{${\bf 0.386 \pm 0.075}$}
\\
\midrule
\multicolumn{8}{c}{MIMIC-III} \\
\midrule
\multirow{5}{*}{\rotatebox[origin=c]{90}{\textit{Unidirectional}}}& 
\multicolumn{1}{l}{GRU-D \cite{che2018}} & 
\multicolumn{1}{c}{$0.583 \pm 0.005$} & 
\multicolumn{1}{c}{$0.926 \pm 0.005$} & 
\multicolumn{1}{c}{$0.973 \pm 0.105$} &
\multicolumn{1}{c}{$0.585 \pm 0.007$} & 
\multicolumn{1}{c}{$0.930 \pm 0.007$} &
\multicolumn{1}{c}{$0.978 \pm 0.096$}
\\ 
& \multicolumn{1}{l}{RITS-I \cite{cao2018}} & 
\multicolumn{1}{c}{$0.457 \pm 0.007$} & 
\multicolumn{1}{c}{$0.725 \pm 0.005$} & 
\multicolumn{1}{c}{$0.809 \pm 0.111$} &
\multicolumn{1}{c}{$0.464 \pm 0.006$} & 
\multicolumn{1}{c}{$0.737 \pm 0.003$} &
\multicolumn{1}{c}{$0.820 \pm 0.099$}
\\
& \multicolumn{1}{l}{RITS \cite{cao2018}} & 
\multicolumn{1}{c}{$0.332 \pm 0.006$} & 
\multicolumn{1}{c}{$0.527 \pm 0.005$} &  
\multicolumn{1}{c}{${\bf 0.562 \pm 0.099}$} &
\multicolumn{1}{c}{$0.356 \pm 0.004$} & 
\multicolumn{1}{c}{$0.567 \pm 0.003$} &
\multicolumn{1}{c}{${\bf 0.604 \pm 0.096}$}
\\
& \multicolumn{1}{l}{\bf V-RIN (Ours)} &
\multicolumn{1}{c}{$0.332 \pm 0.007$} &
\multicolumn{1}{c}{$0.527 \pm 0.008$} &
\multicolumn{1}{c}{$0.617 \pm 0.117$} &
\multicolumn{1}{c}{$0.358 \pm 0.005$} &
\multicolumn{1}{c}{$0.570 \pm 0.003$} &
\multicolumn{1}{c}{$0.636 \pm 0.096$}
\\
& \multicolumn{1}{l}{\bf V-RIN-full (Ours)} &
\multicolumn{1}{c}{${\bf 0.327 \pm 0.005}$} &
\multicolumn{1}{c}{${\bf 0.520 \pm 0.006}$} &
\multicolumn{1}{c}{$0.586 \pm 0.103$} &
\multicolumn{1}{c}{${\bf 0.353 \pm 0.004}$} &
\multicolumn{1}{c}{${\bf 0.561 \pm 0.001}$} &
\multicolumn{1}{c}{$0.610 \pm 0.098$}
\\
\midrule
\multirow{5}{*}{\rotatebox[origin=c]{90}{\textit{Bidirectional}}}& 
\multicolumn{1}{l}{M-RNN \cite{yoon2017}} & 
\multicolumn{1}{c}{$0.441 \pm 0.007$} & 
\multicolumn{1}{c}{$0.700 \pm 0.006$} & 
\multicolumn{1}{c}{$0.758 \pm 0.110$} &
\multicolumn{1}{c}{$0.451 \pm 0.006$} & 
\multicolumn{1}{c}{$0.717 \pm 0.004$} &
\multicolumn{1}{c}{$0.776 \pm 0.096$}
\\
& \multicolumn{1}{l}{BRITS-I \cite{cao2018}} & 
\multicolumn{1}{c}{$0.421 \pm 0.007$} & 
\multicolumn{1}{c}{$0.668 \pm 0.005$} &  
\multicolumn{1}{c}{$0.732 \pm 0.110$} &
\multicolumn{1}{c}{$0.429 \pm 0.005$} & 
\multicolumn{1}{c}{$0.682 \pm 0.002$} &
\multicolumn{1}{c}{$0.745 \pm 0.100$}
\\
& \multicolumn{1}{l}{BRITS \cite{cao2018}} & 
\multicolumn{1}{c}{$0.313 \pm 0.005$} & 
\multicolumn{1}{c}{$0.497 \pm 0.006$} & 
\multicolumn{1}{c}{${\bf 0.530 \pm 0.098}$} &
\multicolumn{1}{c}{${\bf 0.333 \pm 0.004}$} & 
\multicolumn{1}{c}{${\bf 0.529 \pm 0.003}$} &
\multicolumn{1}{c}{${\bf 0.566 \pm 0.095}$}
\\
& \multicolumn{1}{l}{\bf V-RIN (Ours)} &
\multicolumn{1}{c}{$0.314 \pm 0.007$} &
\multicolumn{1}{c}{$0.499 \pm 0.008$} &
\multicolumn{1}{c}{$0.584 \pm 0.115$} &
\multicolumn{1}{c}{$0.339 \pm 0.003$} &
\multicolumn{1}{c}{$0.539 \pm 0.002$} &
\multicolumn{1}{c}{$0.604 \pm 0.099$}
\\
& \multicolumn{1}{l}{\bf V-RIN-full (Ours)} &
\multicolumn{1}{c}{${\bf 0.311 \pm 0.006}$} &
\multicolumn{1}{c}{${\bf 0.494 \pm 0.006}$} &
\multicolumn{1}{c}{$0.561 \pm 0.084$} &
\multicolumn{1}{c}{${\bf 0.333 \pm 0.005}$} &
\multicolumn{1}{c}{$0.530 \pm 0.004$} &
\multicolumn{1}{c}{$0.579 \pm 0.107$}
\\
\bottomrule
\end{tabular}
\end{table*}

\subsection{Classification Result Analysis}
\label{section:experiments_classification}
We presented the experimental results of the in-hospital mortality prediction in comparison with other competing methods in terms of average AUC and AUPRC in {\revise Table \ref{tab:classifcation_task} for both datasets}. We reported the evaluation of both unidirectional and bidirectional models for 10\% and 5\% removal on both datasets. 
For the case of unidirectional models on PhysioNet with 10\% and 5\% removal, our V-RIN model achieved better performance in terms of AUC and AUPRC by a large margin compared to all comparative models even with the RITS, which utilized both the features and temporal relations, sequentially. The highest AUC was obtained by the V-RIN-full in both removal percentage scenarios. As for bidirectional models, all models in the 10\% and 5\% scenarios improved their performance, except for the M-RNN, which achieved lowest results among all competing methods. Although M-RNN employed similar strategies using bidirectional dynamics, it struggles to perform the task properly. Both AUC and AUPRC of M-RNN was lower than GRU-D, despite GRU-D using only the forward directional data. The highest AUC was achieved by our V-RIN-full with $0.8401 \pm 0.0100$ and $0.8422 \pm 0.0070$ for the 10\% and 5\% removal scenarios, respectively. These findings reassure that the utilization of the uncertainty is truly beneficial in estimating the missing values {\revise to better predict the clinical outcomes}.

As for MIMIC-III, we observed quite similar patterns to PhysioNet, with our model outperforming all competing models. The imbalance ratio, missing ratios, and dimensionality of MIMIC-III are much greater than PhysioNet. We found that the performance of V-RIN and V-RIN-full was comparable to some extent. In unidirectional models with 10\%, our V-RIN-full achieved the highest AUC and AUPRC, whereas the V-RIN achieved the highest performance by a small margin in the bidirectional models. This contrasts with the 5\% scenarios, where V-RIN-full with bidirectional models achieved the highest AUC and AUPRC of $0.8705 \pm 0.0148$ and $0.4199 \pm 0.016$, respectively.

\begin{figure*}[t]
\centering
\includegraphics[width=0.97\textwidth]{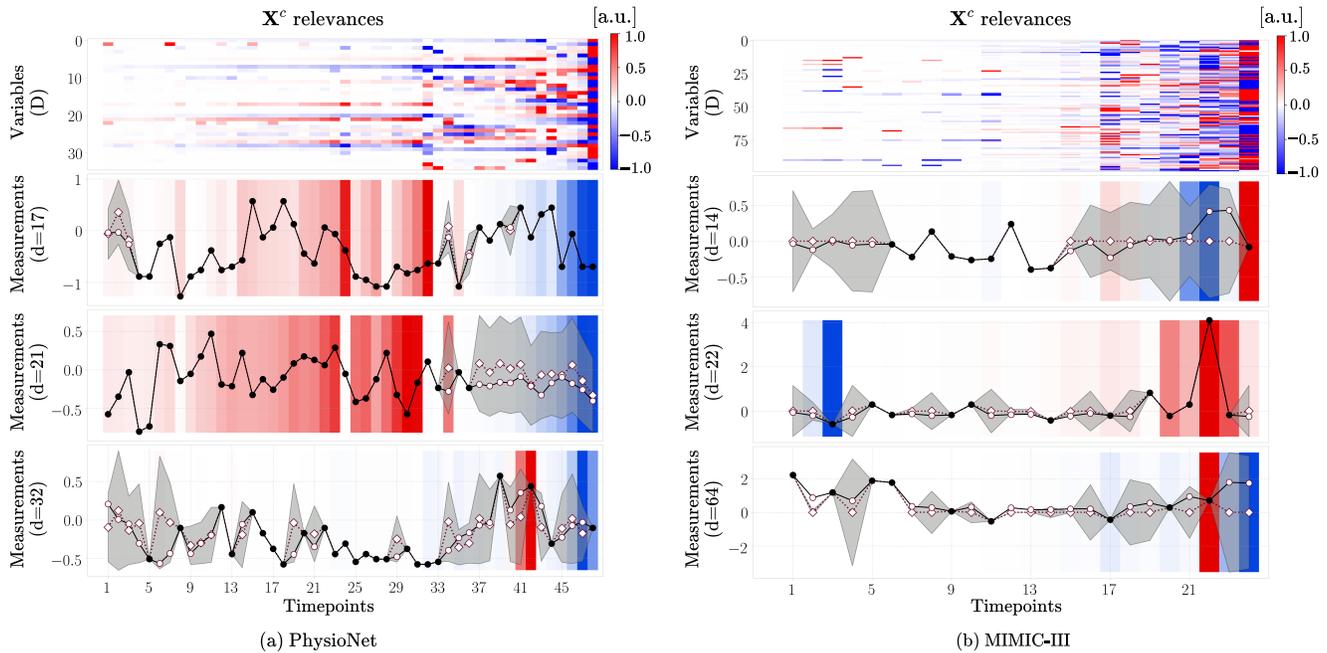}
\caption{Analysis of imputed values relevance in carrying classification task by utilizing LRP \cite{montavon2018,alber2018} to our recurrent imputation network. We obtained the positive and negative relevances depicted as red and blue, respectively, while illustrate the observed values with black circles ($\bullet$), imputed values by VAEs with hollow diamonds ($\diamond$) with shaded areas as its corresponding uncertainties, and combined estimates by means of recurrent imputation network as hollow circles ($\circ$). Color intensity is normalized to the maximum absolute relevance per feature over time.}
\label{fig:lrp_timeseries}
\end{figure*}

\subsection{Imputation Results Analysis}
\label{section:experiments_imputation}
As the secondary task, we further evaluated the imputation performance of our proposed model in contrast with the comparative models in {\revise Table \ref{tab:imputation_task}}. In PhysioNet, GRU-D which exploited the mean and temporal relations in imputing the values struggle to achieve good performance. By using bidirectional dynamics, M-RNN obtained better results than GRU-D. However, those models are still inferior to the RITS variants in both directional scenarios. Overall, both of our proposed models revealed its imputation robustness on this dataset, by exhibiting best performance, in both unidirectional and bidirectional scenario, consistently. {\revise Meanwhile, overall results on MIMIC-III showed that our proposed model performed better than other comparative models on MAE and MRE, and slightly inferior to RITS and BRITS on MSE. Further investigation on these findings are worthy for our future work to uncover the imputation behavior on various clinical datasets with diverse (patients') characteristics.} 


We further exploited layer-wise relevance propagation (LRP) \cite{montavon2018} through a publicly available toolbox: \mbox{the innvestigate} \cite{alber2018} to examine the relevance of the features. Specifically, we employed the LRP-$\alpha_1\beta_0$ rule \cite{montavon2018,alber2018} to discover which features induce the activation of the neurons in carrying out the classification task. As a result, we depicted the imputed values as the input to the recurrent imputation network and further revealed the positive and negative relevance of each feature over time on both datasets in Fig. \ref{fig:lrp_timeseries}. The observed values are illustrated as black circles. By means of VAEs, we obtained the imputation estimates, which are highlighted by the hollow diamond markers alongside the uncertainties, illustrated with the shaded areas. Then, by employing the recurrent imputation network, we acquired the final estimates to the missing values depicted as the hollow circles. Overall, we observed that the relevances were getting stronger toward the end of the time period, regardless of its sign. Furthermore, compared to the observed values, the missing value  estimates with high uncertainties tend to hold a low relevances. Thus, this demonstrates the benefits of the uncertainty utilization in the recurrent imputation networks for the downstream task.

\section{Conclusion}
\label{section:conclusion}
In this study, we proposed a novel unified framework consisting of imputation and prediction networks for sparse high-dimensional multivariate time series. It combined a deep generative model with a recurrent model to capture feature correlations and temporal relations for estimating the missing values by taking into account the uncertainty. We utilized the uncertainties as the fidelity of our estimation and incorporated them for predicting clinical outcomes. We evaluated the effectiveness of the proposed model with the PhysioNet 2012 Challenge and MIMIC-III datasets as the real-world EHR multivariate time series data, proving the superiority of our model in the in-mortality prediction task, compared to other comparative state-of-the-art models in the literature.

\section*{Acknowledgment}
This work was supported by Institute of Information \& communications Technology Planning \& Evaluation(IITP) grant funded by the Korea government(MSIT) (No. 2019-0-00079, \mbox{Department} of Artificial Intelligence (Korea University)) and the National Research Foundation of Korea(NRF) grant funded by the Korea government(MSIT) (No. 2019R1A2C1006543).



\ifCLASSOPTIONcaptionsoff
  \newpage
\fi


\bibliographystyle{IEEEtran}
\bibliography{main}

\end{document}

%% file: math.tex
\usepackage{amsmath,amsfonts,bm}

\def\eqref#1{equation~\ref{#1}}

\def\1{\bm{1}}

\def\rvb{{\mathbf{b}}}
\def\rvc{{\mathbf{c}}}

\def\rvh{{\mathbf{h}}}
\def\rvu{{\mathbf{i}}}

\def\rvm{{\mathbf{m}}}

\def\rvr{{\mathbf{r}}}
\def\rvs{{\mathbf{s}}}

\def\rvu{{\mathbf{u}}}

\def\rvx{{\mathbf{x}}}
\def\rvy{{\mathbf{y}}}
\def\rvz{{\mathbf{z}}}

\def\rmM{{\mathbf{M}}}

\def\rmU{{\mathbf{U}}}
\def\rmV{{\mathbf{V}}}
\def\rmW{{\mathbf{W}}}
\def\rmX{{\mathbf{X}}}

\def\rmZ{{\mathbf{Z}}}

\def\vmu{{\bm{\mu}}}

\def\mSigma{{\bm{\Sigma}}}

\DeclareMathAlphabet{\mathsfit}{\encodingdefault}{\sfdefault}{m}{sl}
\SetMathAlphabet{\mathsfit}{bold}{\encodingdefault}{\sfdefault}{bx}{n}

\def\gD{{\mathcal{D}}}
\def\gE{{\mathcal{E}}}
\def\gF{{\mathcal{F}}}

\def\gL{{\mathcal{L}}}

\newcommand{\mOne}{{\bf 1}}
\newcommand{\mZero}{{\bf 0}}
\newcommand{\mIdentity}{{\bf I}}
\newcommand{\mDelta}{{\mathbf{\Delta}}}
\newcommand{\vdelta}{{\mathbf{\Delta}}}
\newcommand{\vgamma}{{\bm{\gamma}}}
\newcommand{\vsigma}{{\bm{\sigma}}}
\newcommand{\vepsilon}{{\bm{\epsilon}}}
\newcommand{\vupsilon}{{\bm{\upsilon}}}
\newcommand{\vTheta}{{\bm{\Theta}}}
\def\vtau{{\bm{\tau}}}
\newcommand{\withdim}[1]{{\ \in \mathbb{R}^{#1}}}

\newcommand{\KLDiv}[2]{{D_{KL}[#1{\lVert}#2]}}